# Learning to Localize Sound Source in Visual Scenes

Arda Senocak[1]   Tae-Hyun Oh[2]   Junsik Kim[1]   Ming-Hsuan Yang[3]   In So Kweon[1]
Dept. EE, KAIST, South Korea[1]
MIT CSAIL, MA, USA[2]
Dept. EECS, University of California, Merced, CA, USA[3]

## Abstract

*Visual events are usually accompanied by sounds in our daily lives. We pose the question: Can the machine learn the correspondence between visual scene and the sound, and localize the sound source only by observing sound and visual scene pairs like human? In this paper, we propose a novel unsupervised algorithm to address the problem of localizing the sound source in visual scenes. A two-stream network structure which handles each modality, with attention mechanism is developed for sound source localization. Moreover, although our network is formulated within the unsupervised learning framework, it can be extended to a unified architecture with a simple modification for the supervised and semi-supervised learning settings as well. Meanwhile, a new sound source dataset is developed for performance evaluation. Our empirical evaluation shows that the unsupervised method eventually go through false conclusion in some cases. We show that even with a few supervision, false conclusion is able to be corrected and the source of sound in a visual scene can be localized effectively.*

## 1. Introduction

Visual events are typically coherent with sounds and they are integrated. When we see that car is moving, we hear the engine sound at the same time. Sound carries rich information regarding the spatial and temporal cues of the source within a visual scene. As shown in the bottom example of Figure 1, the engine sound suggests where the source may be in the physical world [11]. This implies that sound is not only complementary to the visual information, but also correlated to visual events.

Humans observe tremendous number of combined visual-audio examples and learn the correlation between them throughout their whole life [11] unconsciously. Because of the correlation between the sound and the visual events, humans can understand the object or the event that causes sound and can localize the sound source even without separate education. Naturally, videos and their corresponding sounds also come together in a synchronized way. Given plenty of video and sound clip pairs, can a learning model learn to associate the sound with visual scene to reveal the sound source location without any supervision in a way similar to human perception to localize sound sources in visual scenes?

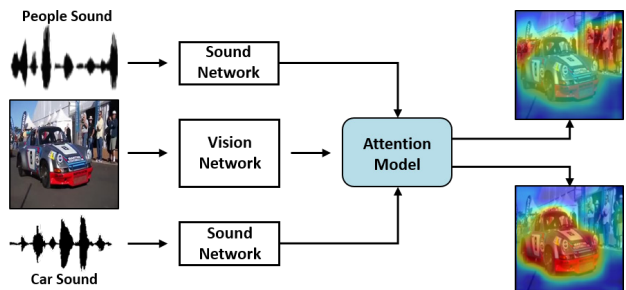

Figure 1. **Where do these sounds come from?** We show an example of interactive sound source localization by the proposed algorithm. In this paper, we demonstrate how to learn to localize the sources (objects) from the sound signals.

In this work, we analyze whether we can model the spatial correspondence between visual and audio information by leveraging the correlation between the modalities based on simply watching and listening to videos in unsupervised way, *i.e.*, learning based sound source localization. We design our model by using two-stream network architecture (sound and visual networks) where each network facilitates each modality and localization module which incorporates the attention mechanism as illustrated in Figure 2.

The learning task for sound source localization from listening is challenging, especially from unlabeled data. Unconstrained videos are likely to contain unrelated audio to the visual contents or audio source that is off the screen, *e.g.*, narration, commenting, *etc*. Another challenge arises from unlabeled information. From our experiments with the proposed unsupervised model, we observe a classical phenomenon [23] in learning theory, *i.e.*, pigeon superstition[1], which biases the resulting localization to be semantically unmatched, in some examples. That is, it is difficult for unsupervised learning methods to disambiguate the sound sources purely based on correlations without certain amount of supervision as illustrated in Figure 3. Comparing to humans, what makes learning mechanisms reaching such false conclusions is always a crucial question. One feasible way to address this problem is to empower the learner with prior knowledge.

---

[1]It is an experiment [26] about delivering food to hungry pigeons in a cage at regular time intervals regardless of the bird behavior. When food was

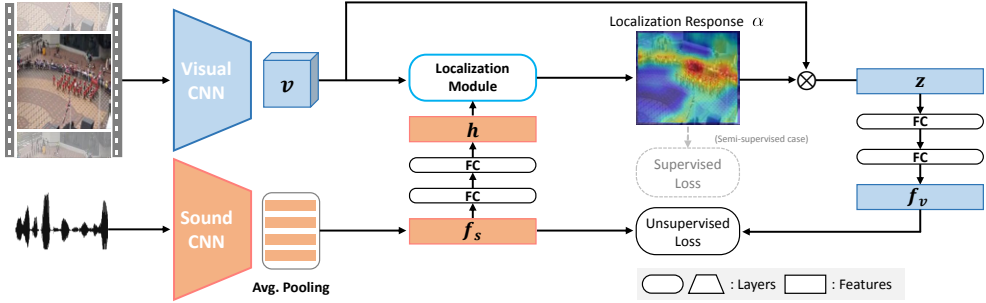

Figure 2. **Network Architecture.** This architecture is designed to tackle the problem of sound source localization with unsupervised learning. The network uses frame and sound pairs to learn to localize sound source. Each modality is processed in its own network. After integrating(correlating) the information from the sound context vector **h** and the activations of visual network, attention mechanism localizes the sound source. By adding supervised loss component into this architecture, it can be converted to a unified architecture which can work as supervised or semi-supervised learning as well.

We correct this issue by providing some prior knowledge in a semi-supervised setting. By adding a supervised loss to our network, we propose a unified architecture that is able to learn within the unsupervised or semi-supervised framework based on whether annotated data is available or not. To incorporate in our unified architecture and also to evaluate our proposed methods, we annotate a new sound source localization dataset. To the best of our knowledge, our dataset is the first to address the problem of learning based sound localization. The main contributions of this work are summarized as follows:

• We introduce a learning framework to localize sound source using the attention mechanism, which is guided by sound information, with a paired sound and video frame. Thus, the sound source localization can be interactive with given sound input.

• We propose a unified end-to-end deep convolutional neural network architecture that accommodates unsupervised, semi-supervised, and fully-supervised learning.

• We collect and annotate a new sound source localization dataset, which provides supervised information and facilitates quantitative and qualitative analysis.

---

first delivered, it is found that each pigeon was engaging in some activity. Then they started doing the same action, believing that by acting in that way food would arrive, *i.e.*, reinforced to do a specific action. Such self-reinforcement occurs regardless of trueness of causality of the event and its chance. Some of such fundamental issues that naturally occur in the context of animal learning also appear in machine learning [23].

## 2. Related Work and Problem Context

Cross-modality signals have been used as supervised information for numerous tasks [20, 15, 10]. The recent years have witnessed significant progress in understanding the correlation between sound and visual signals in videos [2, 21, 20, 27, 3, 1]. To put this work into proper context, we review recent methods on joint sound-visual models, sound source localization, and attention mechanisms.

Recent methods [21, 20] consider sound as supervisory signal by virtue of its natural synchronization with visual input, while the SoundNet [2] regards visual imagery as supervision for sound. These approaches leverage the correlation as well as co-occurrence of the two modalities, and learn representation of one the modalities while using the other one. We use a network similar to the SoundNet as our audio module in this work. Concurrently, Aytar *et al.* [3] and Arandjelović *et al.* [1] propose to learn jointly aligned cross-modal representations. We note while Arandjelović *et al.* shows the activation maps that localize objects, the localization results are from solely examining the units in the vision subnetwork. This localization method corresponds to visual saliency which is not interactively estimated according to the given sound. In contrast, our networks have a bridge layer that interacts between the two modalities and reveals the localization information of the sound source.

Our work is motivated by the findings in psychology and cognitive science [11, 16, 19, 24, 6, 22] on sound source localization capability of humans. Gaver *et al.* [11] study how humans learn about objects and events from sound in everyday listening. This study elucidates how humans can find the relationship between visual and sound domains in an event centric view. Numerous methods in this line of work analyze the relationship between visual information and sound localization. The findings demonstrate that visual information correlated to sound improves the efficiency of search [16] and accuracy of localization [24]. Recent research [19, 6, 22] extends the findings of human perfor-mance on sound source localization against visual information in 3D space. These studies evidently show that sound source localization capability of humans is guided by visual information, and two sources of information are closely correlated that humans can unconsciously learn the capability.

Prior to the recent advances of deep learning, computational methods for sound source localization rely on synchrony [12] of low-level features of sounds and videos (*e.g.*, raw waveform signals and intensity values respectively), spatial sparsity prior of audio-visual events [18], low-dimensionality [9], and hand-crafted motion cues as well as segmentation [5, 14]. In contrast, the proposed network is developed in an unsupervised manner by only watching and listening to videos without using any manually-designed constraints such as motion. In addition, our semi-supervised architecture accommodates minimal prior information for better performance.

Acoustic based approach [29, 33] has been practically used in surveillance and instrumentation engineering. It requires specific devices, *e.g.*, microphone arrays, to capture phase differences of sound arrival. In this work, we learn sound source localization in the visual domain without any special devices but a microphone to capture sound.

Inspired by human attention mechanism [7], computational visual attention models [30, 32] have been developed. However not only visual attentions but also the sound localization behavior in imagery resemble to the human attention. In this work, we adopt the same attention mechanism philosophy [30] to allow our networks to interact with sound context and visual representation across spatial axes.

## 3. Proposed Algorithm

We design a neural network to address the problem of vision based sound localization within the unsupervised learning framework. In order to deal with cross-modality signals from sounds and videos, we use a two-stream network architecture. The network composes of three main modules: sound network, visual network and attention model as illustrated in Figure 2. We describe the main components of each module in this section, and present more network details as well as hyper-parameters in the supplementary material.

### 3.1. Sound Network

For sound localization, it is important to capture the concept of sound rather than catching low-level signal [11]. In addition, sound signal is a 1-D signal with varying temporal length. We represent sound signals by using the convolutional module (conv), rectified linear unit (ReLu) and pooling (pool), and stacking layers to encode high-level concepts [31].

We use a 1-D deep convolutional architecture which is invariant to input length as fully convolutional networks via the use of average pooling over sliding windows. The proposed sound network consists of 10 layers and takes raw waveform as input. The first conv layers (up to conv8) are similar to the SoundNet [2], but with 1000 filters followed by average pooling across the temporal axis within a sliding window (*e.g.*, 20 seconds in this work).

Similar to the global average pooling which can handle variable length inputs to be a fixed dimension vector [28], the output activation of conv8 followed by average pooling is always a single 1000-D vector by a sliding window. We denote the sound representation after average pooling as $\mathbf{f}_s$.

To encode higher level concept of the sound signals, the 9-th and 10-th layers consist of ReLU followed by fully connected (FC) layers. The output of the 10-th FC layer (FC10) is 512-D, and is denoted as $\mathbf{h}$. We use $\mathbf{h}$ to interact with features from the visual network, and enforce $\mathbf{h}$ to resemble visual concepts. Among these two features, we note $\mathbf{f}_s$ preserves more sound concept while $\mathbf{h}$ captures correlation information related to visual signals.

### 3.2. Visual Network

The visual network is composed of the image feature extractor and localization module. To extract features from visual signals, we use an architecture similar to the VGG-16 model [25] up to conv5_3 and feed a color video frames of size $H \times W$ as input. We denote the activation of conv5_3 as $\mathcal{V} \in \mathbb{R}^{H' \times W' \times D}$, where $H'=\lfloor \frac{H}{16} \rfloor$, $W'=\lfloor \frac{W}{16} \rfloor$ and $D = 512$. Each 512-D activation vector from conv5_3 contains local visual context information, and spatial information is preserved in $H' \times W'$ grid.

We make the activation $\mathcal{V}$ interact with the sound embedding $\mathbf{h}$ for revealing sound source location information in the grid, which is denoted as the localization module. This localization module returns a confidence map of sound source and a representative visual feature vector $\mathbf{z}$ corresponding to location of source of the given input sound. Once we obtain the visual feature $\mathbf{z}$, it goes through two {ReLu-FC} blocks to compute the visual embedding $\mathbf{f}_v$, which is the final output of the visual network.

### 3.3. Localization Network

Given extracted visual and sound concepts, the localization networks generate the sound source location. We compute a soft confidence score map as a sound source location representation. This may be modeled based on the attention mechanism in the human visual system [7], where according to given conditional information, related salient features are dynamically and selectively brought out to the foreground. This motivates us to exploit the neural attention mechanism [30, 4] in our context.

For simplicity, instead of using a tensor representation for the visual activation $\mathcal{V} \in \mathbb{R}^{H' \times W' \times D}$, we denote the visual activation as a reshaped matrix form $\mathbf{V} =$ $[\mathbf{v}_1;\cdots;\mathbf{v}_M]\in\mathbb{R}^{M\times D}$, where $M = H'W'$. For each location $i \in \{1,\cdots,M\}$, the attention mechanism $g_{\tt att}$ generates the positive weight $\alpha_i$ by the interaction between the given sound embedding $\mathbf{h}$ and $\mathbf{v}_i$, where $\alpha_i$ is the attention measure. The attention $\alpha_i$ can be interpreted as the probability that the grid $i$ is likely to be the right location related to the sound context, and computed by

$$\alpha_i = \frac{\exp(a_i)}{\sum_j \exp(a_j)}, \quad \text{where } a_i = g_{\tt att}(\mathbf{v}_i, \mathbf{h}), \qquad (1)$$

where the normalization by the softmax is suggested by [4].

In contrast to the work [30, 4] that uses a multi-layer perceptron as $g_{\tt att}$, we use the simple normalized inner product operation that does not require any learning parameters. Furthermore, it is intuitively interpretable that the operation measures the cosine similarity between two heterogeneous vectors, $\mathbf{v}_i$ and $\mathbf{h}$, *i.e.*, correlation. We also propose another alternative attention mechanism to suppress negative correlation values: The two mechanisms are defined as

$$[\text{Mechanism 1}] \quad g_{\tt cos}(\mathbf{v}_i, \mathbf{h}) = \bar{\mathbf{v}}_i^\top \bar{\mathbf{h}}, \qquad (2)$$
$$[\text{Mechanism 2}] \quad g_{\tt ReLu}(\mathbf{v}_i, \mathbf{h}) = \max(\bar{\mathbf{v}}_i^\top \bar{\mathbf{h}}, 0), \qquad (3)$$

where $\bar{\mathbf{x}}$ denotes a $\ell_2$-normalized vector. This is different from the mechanism proposed in [30, 4, 32]. Zhou *et al.* [32] use a typical linear combination without normalization, and thus it can have an arbitrary range of values. Both mechanisms in this work are based on the cosine similarity of the range $[-1, 1]$.

The attention measure $\boldsymbol{\alpha}$ computed by either mechanism describes the sound and visual context interaction in a map. To give a connection to $\boldsymbol{\alpha}$ with sound source location, similar to [30, 4], we further process to compute the representative context vector $\mathbf{z}$ that corresponds to the local visual feature at the sound source location.

Assuming that $\mathbf{z}$ is a stochastic random variable and $\boldsymbol{\alpha}$ represents the sound source location reasonably well, *i.e.*, follows $p(i|\mathbf{h})$, the visual feature $\mathbf{z}$ can be obtained by

$$\mathbf{z} = \mathbb{E}_{p(i|\mathbf{h})}[\hat{z}] = \sum_{i=1}^M \alpha_i \mathbf{v}_i. \qquad (4)$$

As described in Section 3.2, we transform a visual feature vector $\mathbf{z}$ to a visual representation $\mathbf{f}_v$. We adapt $\mathbf{f}_v$ to be comparable with the sound features $\mathbf{f}_s$ obtained from the sound network, such that we learn the features to share embedding space. During the learning phase, the error back-propagation encourages $\mathbf{z}$ to be related to the sound context. However, since the only sound context comes from $\boldsymbol{\alpha}$, the attention measure $\boldsymbol{\alpha}$ is learned to localize the sound. We present details on learning to localize sound in Section 4.

## 4. Localizing Sound Source via Listening

Our learning model determines a video frame and audio signals are similar to each other or not at each spatial location. With the proposed two-stream network, we obtain predictions from each subnetwork for the frame and the sound. If the visual network considers a given frame contains a motorcycle and sound network also returns similar output, then the predictions of these two networks are likely to be similar and close to each other in the feature space, and vice versa. This provides valuable information for learning to localize sound sources in different settings.

**Unsupervised learning.** For this setting, we use the features $\mathbf{f}_v$ and $\mathbf{f}_s$ from two networks. In the feature space, We impose that $\mathbf{f}_v$ and $\mathbf{f}_s$ from the corresponding pairs (positive) are close to each other, while non-corresponding (negative) pairs are far from each other.

We use $\mathbf{f}_v$ from a video frame as a query, and its positive pairs are obtained by taking $\mathbf{f}_s$ from the sound wave from a sliding window around the frame of the same video, while negative ones are extracted from another random video. Given these positive and negative pairs, we use the triplet loss [13] for learning. The triplet input is composed of query, positive sample and negative sample. The loss is designed to map the positive samples into the same location with the query in the feature space, while mapping the negative samples into distant locations.

A triplet network outputs two distance terms:

$$T(\mathbf{f}_v, \mathbf{f}_s^-, \mathbf{f}_s^+) = [\|\mathbf{f}_v - \mathbf{f}_s^+\|_2, \|\mathbf{f}_v - \mathbf{f}_s^-\|_2] = [d_+, d_-],$$

where $T(\cdot)$ denotes the triplet network, $(\mathbf{x}, \mathbf{x}_+, \mathbf{x}_-)$ denotes a triplet of query, positive and negative sample. To impose the constraint $d_+ < d_-$, we use the distance ratio loss [13]. The unsupervised loss function is defined as

$$\mathcal{L}_U(D_+, D_-) = \|D_+\|_2^2 + \|1 - D_-\|_2^2, \qquad (5)$$

where $D_\pm = \frac{\exp(d_\pm)}{\exp(d_+)+\exp(d_-)}$.

For the positive pair, the unsupervised loss imposes the visual feature $\mathbf{f}_v$ to be resembled to $\mathbf{f}_s$. In order for $\mathbf{z}$ to generate such $\mathbf{f}_v$, the weight $\boldsymbol{\alpha}$ needs to select causal locations by the correlation between $\mathbf{h}$ and $\mathbf{v}$. This results in $\mathbf{h}$ to share the embedding space with $\mathbf{v}$, and $\mathbf{f}_s$ also needs to encode the context information that correlated with video frame. This forms a cyclic loop as shown in Figure 2, which allows to learn a shared representation that can be used for sound localization.

Although this unsupervised learning method performs well, we encounter the pigeon superstition issue as discussed in Section 1. For example, as shown in Figure 3, even though we present a train sound with a train image, the proposed model localizes railways rather than the train. This is a false conclusion by the model that causes some semantically unmatched results.

These results can be explained as follows. In the early stage, our model mistakenly concludes with false random output (*e.g.*, activation on the road given car sound). However, it obtains a good score (as it is paired), thereby the model is trained to behave similarly for such scenes. It again reinforces its model to receive good scores in similar examples. Additionally, in the case of the road example, the proposed network consistently sees road with similar car sound during training. Since road has simpler appearance and typically occupies larger regions compared to diverse appearance of the car (or non-existence of the car in the frame), it is difficult for the model to discover the true causality relationship with the car without supervisory feedbacks. This ends up biasing toward a certain semantically unrelated output, as in pigeon superstition problem.

As a simple remedy to this issue, we provide some prior knowledge with supervisory signals in the semi-supervised setting and algorithm learns successfully (see the last column of Figure 3).

**Semi-supervised learning.** Even a small amount of prior knowledge may induce better inductive bias [23]. We add a supervised loss into the proposed network architecture under the semi-supervised learning setting. To this end, we formulate the semi-supervised loss,

$$\mathcal{L}(\mathbf{f}_v, \mathbf{f}_s^+, \mathbf{f}_s^-, \boldsymbol{\alpha}, \boldsymbol{\alpha}_{\text{GT}}) = \\ \mathcal{L}_U(\mathbf{f}_v, \mathbf{f}_s^+, \mathbf{f}_s^-) + \lambda(\boldsymbol{\alpha}_{\text{GT}}) \cdot \mathcal{L}_S(\boldsymbol{\alpha}, \boldsymbol{\alpha}_{\text{GT}}), \quad (6)$$

where $\mathcal{L}_U$ and $\mathcal{L}_S$ denote unsupervised and supervised losses respectively, $\boldsymbol{\alpha}_{\text{GT}}$ denotes the ground-truth (or reference) attention map, and $\lambda(\cdot)$ is a function for controlling the data supervision type. The unsupervised loss $\mathcal{L}_U$ is same as (5). The supervised loss $\mathcal{L}_S$ can be the mean square error or cross entropy loss. Empirically, the proposed model with the cross entropy loss performs slightly better. The adopted cross entropy loss is defined by

$$\mathcal{L}_S(\boldsymbol{\alpha}, \boldsymbol{\alpha}_{\text{GT}}) = -\sum_i \alpha_{\text{GT},i} \log(\alpha_i), \quad (7)$$

where $i$ denotes the location index of the attention map and $\alpha_{\text{GT},i}$ is a binary value. We set $\lambda(\mathbf{x}) = 0$ if $\mathbf{x} \in \emptyset$, 1 otherwise. With this formulation, we can easily adapt the loss to be either supervised or unsupervised one according to the existence of $\boldsymbol{\alpha}_{\text{GT}}$ for each sample. In addition, (6) can be directly utilized for the fully supervised training.

## 5. Experimental Results

For evaluation, we first construct a new sound source localization dataset which facilitates quantitative and qualitative evaluation. Based on this dataset, we evaluate the capability of the proposed method with various settings. As aforementioned, while the proposed network is learned to interact with given sound, the semantically unmatched localized results has been observed when we tackle the problem in a unsupervised way. In this section, we discuss our empirical observations, and demonstrate how such issues can be corrected with a few supervision. In addition, we compare our unified network in unsupervised, semi-supervised and supervised learning schemes. More results can be found in the supplementary material. The source code and dataset will be made available to the public.

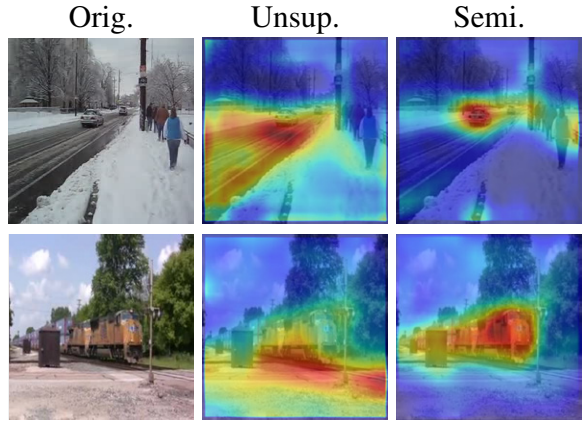

Figure 3. **Semantically unmatched results.** We show some of the cases where proposed network draws false conclusions. We correct this issue by providing a prior knowledge.

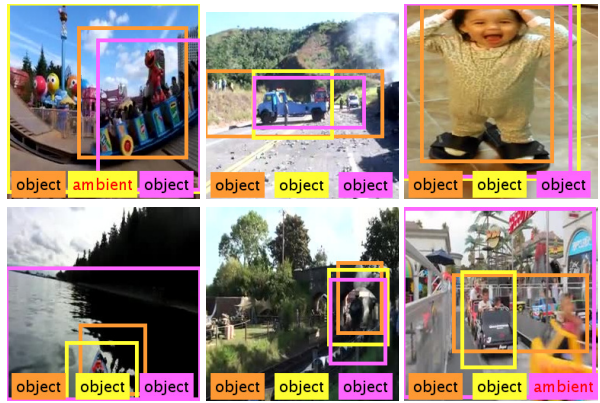

Figure 4. **Sound Source Localization Dataset.** Human annotators annotated the location of the sound source and the type of the source (object vs. non-object/ambient). This dataset is used for testing how well our network learned the sound localization and also for providing a supervision to unified architecture.

### 5.1. Dataset

In order to train our network to localize the sound sources, we leverage the large unlabeled Flickr-SoundNet [2] dataset, which consists of more than two million unconstrained sound and image pairs. We use a random subset of 144k pairs to train our network.

For benchmark purpose, we collect a new dataset that source of sounds are annotated in image coordinates using pairs from the Flickr-SoundNet set. This dataset facilitates not only quantitative and qualitative evaluation, but also provides annotations for training supervision models. We randomly sample 5k frames and its corresponding sound from Flickr-SoundNet dataset. We make three subjects an-notate sound source locations by giving a generic instructions as follows: 1) listen 20 secs. of sound and draw bounding box on the frame at the regions where the sound would come from, and 2) tag the bounding box as object or ambient.

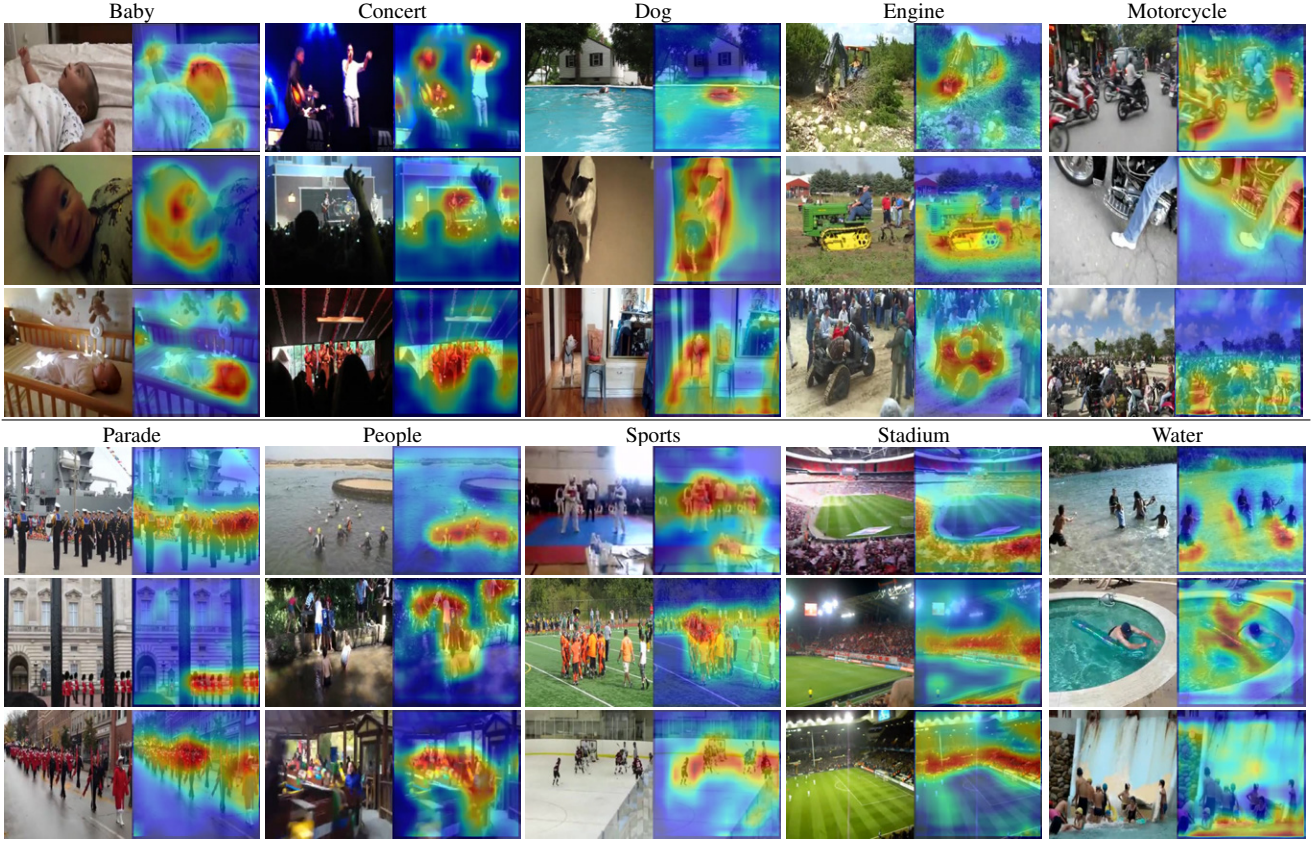

Figure 5. **Qualitative Sound Localization Results from Unsupervised Network.** We visualize some of the sound source locations. We feed image and sound pairs through our unsupervised network and it highlights the regions that sound is originated. Titles of the columns are subject and shown only for visualization purpose to give an idea about the sound context to readers: We do not use explicit labels.

Since the dataset we use contains unconstrained videos, some frames do not have the sound source in the frame or sound source cannot be represented by drawing a bounding box, *e.g.*, wind sound. The tag is used to distinguish this case as "object" or "ambient/not object" for each bounding box. After annotation process, we filter out "ambient/not object" image-sound pairs. Among remaining pairs, we select the ones that all subjects agree that sound indicating objects present in the frame. From the set of 2,786 pairs, we randomly sample 250 to construct a testing set and use the rest for training. Figure 4 shows some sample images.

### 5.2. Results and Analysis

We introduce a metric for quantitative performance evaluation of sound localization.

**Evaluation metrics.** We have three annotations from three respective subjects for each data. As some examples could be ambiguous, *e.g.*, the left and right examples in the bottom row of Figure 4, we present a consensus metric, *i.e.*, consensus intersection over union (cIoU), to take issue of multiple annotations into account. Similar to the consensus metric in the context of the VQA task [17], we assign scores to each pixel according to consensus of multiple annotations.

First, we convert the bounding box annotations to binary maps $\{\mathbf{b}_j\}_{j=1}^N$, where $N$ is the number of subjects. We extract a representative score map $\mathbf{g}$ by collapsing $\{\mathbf{b}_j\}$ across subjects but with considering consensus as

$$\mathbf{g} = \min\left(\sum_{j=1}^N \frac{\mathbf{b}_j}{\#\text{consensus}}, 1\right), \quad (8)$$

where $\#\text{consensus} \leq N$ is the parameter means the minimum number of opinion to reach agreement. For example, for a pixel, when positives more than or equal to #consensus are given in the binary map, then the pixel of $\mathbf{g}$ is set to a full score, *i.e.*, 1, otherwise a proportional score. Since we have three subjects, by majority rule, we set #consensus=2 in our experiment. Given this weighted score map $\mathbf{g}$ and predicted location response $\boldsymbol{\alpha}$, we define the cIoU as

$$\text{cIoU}(\tau) = \frac{\sum_{i \in \mathcal{A}(\tau)} g_i}{\sum_i g_i + \sum_{i \in \mathcal{A}(\tau) - \mathcal{G}} 1}, \quad (9)$$

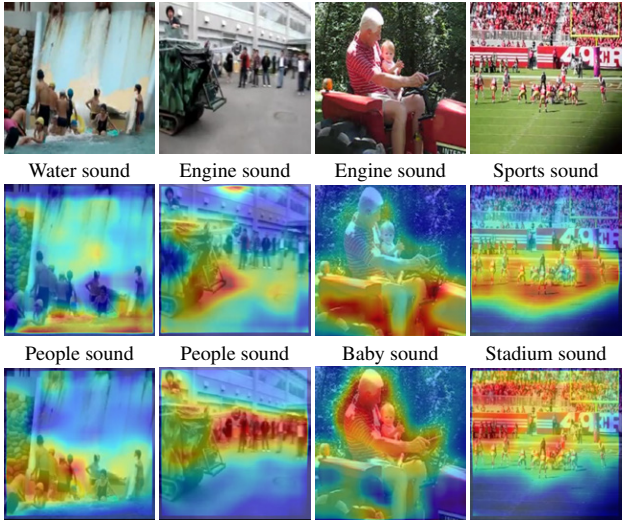

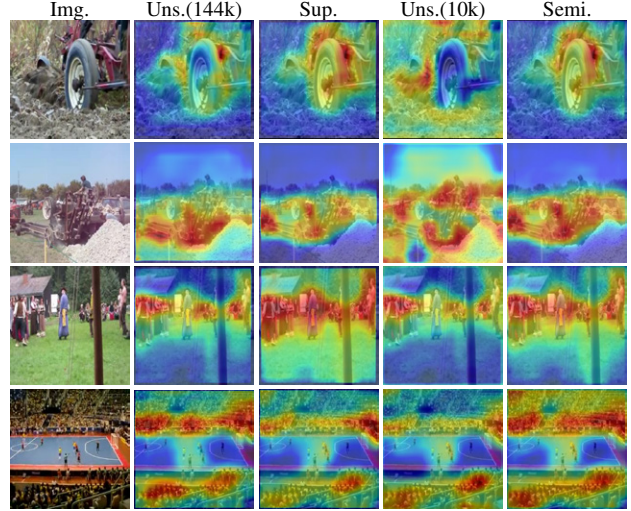

Figure 6. **Interactive sound source localization.** We visualize the responses of the network to different sounds while keeping the frame same. These results show that our network can localize the source of the given sound interactively.

where $i$ indicates the pixel index of the map, $\tau$ denotes the threshold to judge positiveness, $\mathcal{A}(\tau) = \{i|\alpha_i > \tau\}$, and $\mathcal{G} = \{i|g_i > 0\}$. In (9), $\mathcal{A}$ is the set of pixels with attention intensity higher than threshold, and $\mathcal{G}$ is the set of pixels classified as positives in weighted ground truth. The denominator implies a weighted version of union and the threshold $\tau$ decides the foreground area. Using the common practice in object detection [8], we use 0.5 for the cIoU threshold in the experiments.

**Qualitative Analysis.** In qualitative comparisons, we mainly visualize the localization response $\alpha$. Figure 5 shows the localization results of the image-sound pairs from the Flickr-SoundNet dataset [2] using the proposed *unsupervised learning* approach. Our network learns to localize sound sources on a variety of categories without any supervision. It is interesting to note that sound sources are successfully localized in spite of clutters, and unrelated areas are isolated, *e.g.*, in the "water" column of the Figure 5, people are isolated from the highlighted water areas. As seen in the "concert" examples; scenes include both stage people and the audiences. Even though they have similar appearance, the learned model is still able to distinguish people on the stage from the audiences.

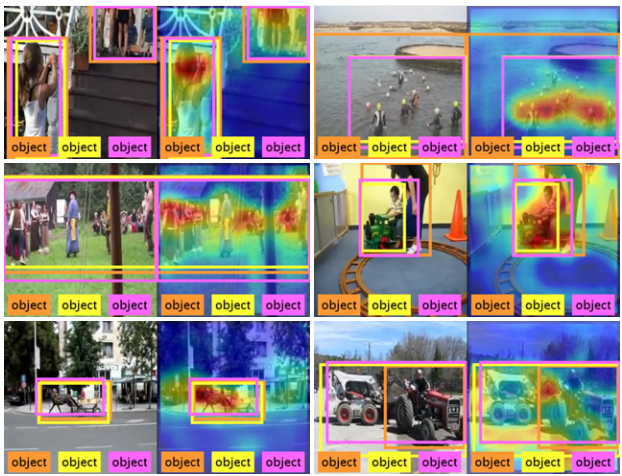

Figure 7. **How well can our network localize the sound sources compare to humans?** Qualitative comparison of localization between our network and human annotations.

Figure 8. **Qualitative Sound Localization Results from Different Learning Methods.** We visualize sound localization results from different learning methods. The supervised method generally localizes the sound source precisely due to the guidance of ground truths. Despite using less supervised data, the semi-supervised approach also gives comparably accurate localization results.

At the first glance, the results may look like hallucinating salient areas or detecting objects regardless of sound contexts. It should be noted that, our network responds interactively according to given sound. Figure 6 shows examples of different input sounds for same images where the localization responses change according to the given sound context. For a frame that contains water and people, when a water sound is given, the water area is highlighted. Similarly, the area containing people is highlighted when the sound source is from humans.

With the network trained in unsupervised way, we qualitatively compare the localization performance with respect to human annotations. Figure 7 shows sample qualitative results where the learned model performs consistently with human perception even though no prior knowledge is used.

While the network learns to localize sound sources in variety of categories without supervision, as aforementioned in Figure 3, there are numerous cases that the unsupervised network falsely concludes the matching between visual and sound contexts. Using the semi-supervised scheme within the unified network model, we can transfer human knowledge to remedy the pigeon superstition issue. Figure 8 shows the results by other learning methods. As expected, supervised learning methods localize objects more precisely with the ground truth super-visionary signals. We note that the proposed semi-supervised model achieves promising results by incorporating supervised and unsupervised data.

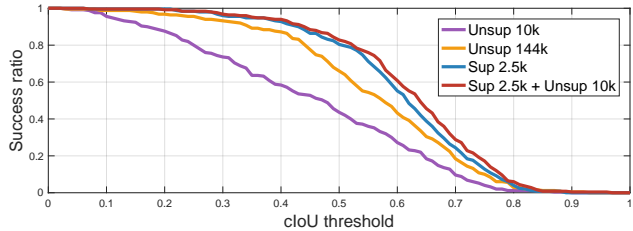

Figure 9. **Success ratio using varying cIoU threshold.** The attention mechanism with softmax without ReLU is used.

Table 1. **Performance evaluation with different learning schemes.** cIoU measures ratio of success samples at 0.5 threshold. AUC measures the area under the graph plotted by varying cIoU threshold from 0 to 1.

|  | softmax | | ReLU+softmax | |
| --- | --- | --- | --- | --- |
|  | cIoU | AUC | cIoU | AUC |
| Unsup. 10k | 43.6 | 44.9 | – | – |
| Unsup. 144k | 66.0 | 55.8 | 52.4 | 51.2 |
| Sup. 2.5k | 80.4 | 60.3 | 82.0 | 60.7 |
| Sup. 2.5k + Unsup. 10k | 82.8 | 62.0 | 84.0 | 61.9 |

| Random | cIoU | AUC |
| --- | --- | --- |
|  | $0.12 \pm 0.2$ | $32.3 \pm 0.1$ |

**Quantitative results.** Table 1 shows the evaluation results using different learning schemes and number of samples. The baseline model is trained in an unsupervised manner with 10k samples. We report random prediction results with 100 runs for reference. The results show that the unsupervised model with 10k samples learns meaningful knowledge from the sound and video pairs. We observe significant improvement when the unsupervised network is trained with a larger number of samples, *i.e.*, 144k samples.

We show the supervised learning results with 2.5k samples as reference. Even the number of samples is smaller than unsupervised method, the learned model performs well. When we train the network in the semi-supervised setting with both supervised and unsupervised loss, the model achieves the best performance. The results suggest there is complementary information from unlabeled data, which facilitates the model generalize well. We plot the success rate of the test samples according to cIoU threshold in Figure 9.

We measure the effect of the number of labeled samples in the semi-supervised scenario in Table 2. The results show that near 1k supervised samples are sufficient for the semi-supervised model to learn well. We note that the proposed model benefits more from combination of both types of data than simply increasing the number of supervised samples.

Table 2. **Semi-supervised learning performance with a different number of supervised samples.**

|  | softmax | | ReLU+softmax | |
| --- | --- | --- | --- | --- |
|  | cIoU | AUC | cIoU | AUC |
| Unsup. 10k | 43.6 | 44.9 | – | – |
| Unsup. 144k | 66.0 | 55.8 | 52.4 | 51.2 |
| Sup. 0.5k + Unsup. 10k | 78.0 | 60.5 | 79.2 | 60.3 |
| Sup. 1.0k + Unsup. 10k | 82.4 | 61.1 | 82.4 | 61.1 |
| Sup. 1.5k + Unsup. 10k | 82.0 | 61.3 | 82.8 | 61.8 |
| Sup. 2.0k + Unsup. 10k | 82.0 | 61.5 | 82.4 | 61.4 |
| Sup. 2.5k + Unsup. 10k | 82.8 | 62.0 | 84.0 | 61.9 |

Table 3. **Performance measure against individual subjects.**

| Subject | Unsup. 144k | | Sup. | | Semi-sup. | |
| --- | --- | --- | --- | --- | --- | --- |
|  | IoU | AUC | IoU | AUC | IoU | AUC |
| Subj. 1 | 58.4 | 52.2 | 70.8 | 55.6 | 74.8 | 57.1 |
| Subj. 2 | 58.4 | 52.4 | 72.0 | 55.6 | 73.6 | 57.2 |
| Subj. 3 | 63.6 | 52.6 | 74.8 | 55.6 | 77.2 | 57.3 |
| Avg. | 60.1 | 52.4 | 72.5 | 55.6 | 75.2 | 57.2 |

We report the IoU performance of each annotator in Table 3 that we can see the human factor in this task. While the numbers across subjects vary slightly, the variance is in a small range. Despite the ambiguity nature of the localization task, the results show that human perception of sound localization in images is consistent.

## 6. Discussion and Conclusion

We tackle a new problem, learning based sound source localization in visual scenes, and build its new benchmark dataset. By empirically demonstrating the capability of our unsupervised network, we show the model plausibly works in a variety of categories but partially, in that the network can often get to false conclusion without prior knowledge. We also show that leveraging small amount of human knowledge can discipline the model, so that it can correct to capture semantically meaningful relationships. These may imply that, by the definition of learnability [23], the task is not fully learnable problem only with unsupervised data, but it can be fixed by providing even small amount of supervision.

The results and conclusion made in this work may allow us to deduce the way of machine understanding about sound source localization in visual scenes. As a likely case, in unsupervised representation learning from sound-video pairs [21, 1], our results may indicate that some of representations behave like the pigeons (as in the second row of "Railway" column in Figure 5 of [1]), and suggest that at least small amount of supervision should be incorporated for sound based representation learning. Additionally, this work would open many potential directions for future research, *i.e.*, multi-modal retrieval, sound based saliency or representation learning and its applications.

**Acknowledgements:** We thank to Wei-Sheng Lai for helpful discussions. We are grateful to the annotators for localizing the sound sources on our dataset.